%% file: main.tex
\documentclass[11pt]{article}

\usepackage[preprint]{acl}

\usepackage{times}
\usepackage{latexsym}
\usepackage[T1]{fontenc}
\usepackage[utf8]{inputenc}
\usepackage{microtype}
\usepackage{inconsolata}
\usepackage{graphicx}
\usepackage{colortbl}
\usepackage{tikz}
\usetikzlibrary{positioning,arrows.meta}
\usepackage[most]{tcolorbox}
\usepackage{amssymb}
\usepackage{compat}

\title{LazyTrain: Limited-resource Allocation toward Zero-waste Yield Optimization in Large Language Model Training}

\author{
\bf Xiaojun Wu$^{\ast\ 1,2}$,
Cehao Yang$^{\ast\ 1,2}$,
Honghao Liu$^{\ast\ 1,2}$,
Xueyuan Lin$^{\ast\ 2}$, \\
\bf Xuhui Jiang$^{1,3}$,
Chengjin Xu$^{1,3}$,
Jia Li$^{\dagger\ 2}$,
Jian Guo$^{\dagger\ 1}$
\\
\\
$^{1}$IDEA Research \\
$^{2}$The Hong Kong University of Science and Technology (Guangzhou) \\
$^{3}$DataArcTech Ltd. \\
}

\newcommand{\method}{\mbox{LazyTrain}\xspace}
\newcommand{\megatrain}{MegaTrain\xspace}

\begin{document}
\maketitle
{
  \renewcommand{\thefootnote}{\fnsymbol{footnote}}
  \footnotetext[1]{Equal contribution.}
  \footnotetext[2]{Corresponding authors.}
}

\begin{abstract}
Training large language models on limited hardware is increasingly a scheduling problem across GPU compute, host memory, PCIe transfer, and storage bandwidth.
Existing offloading systems reduce GPU residency, and \megatrain shows that a CPU-master layer-streaming executor can train large models on a single GPU, but fixed checkpointing and placement heuristics still leave communication exposed on the critical path.
We propose \method, an optimization layer over a layer-streaming executor.
\method formulates checkpoint selection, activation placement, recomputation, and CPU-GPU-NVMe communication overlap as a mixed-integer scheduling problem, then executes the solved policy during training.
It further couples 8-bit optimizer states with fast gradient clipping as a
single Hybrid 8-bit operator: state compression reduces optimizer-state memory,
while fast clipping counteracts the additional CPU-side update overhead.
Across H800 experiments from Qwen2.5-3B to Qwen3.6-27B, \method improves sustained TFLOPS over matched baselines runs by approximately 1.24$\times$;
RTX 3090 experiments likewise increase the maximum feasible batch size by one at each model scale. In the primary Qwen3.6-27B H800 MetaMathQA run, \method reaches 219.95 TFLOPS and 1361 tokens/s at batch size 72, peaks at 68.84\,GB of GPU memory, and obtains 95.42\% exact-match accuracy on the full evaluation split.
The source code is available at \url{https://github.com/DataArcTech/LazyTrain}.
\end{abstract}

\input{secs/01_introduction.tex}
\input{secs/02_related.tex}
\input{secs/03_method.tex}
\input{secs/04_experiments.tex}
\input{secs/09_conclusion.tex}

\clearpage
\bibliography{main}

\clearpage
\input{appx.tex}

\end{document}

%% file: secs/01_introduction.tex
\section{Introduction}

Large language model training increasingly depends on how well a system uses
the full memory hierarchy rather than GPU memory alone. Model parameters,
gradients, optimizer states, and activations can exceed the memory of a single
accelerator even for supervised fine-tuning, while CPU DRAM and local storage
offer much larger capacity at substantially lower bandwidth. Efficient training
therefore requires a schedule that decides not only what to offload, but also
when each transfer can be hidden under GPU computation~\citep{duan2024efficient}.

Existing memory-efficient training systems address important pieces of this
problem. ZeRO-style sharding and offloading reduce device residency for model
states~\citep{rajbhandari2020zero,rajbhandari2021zero}, FSDP and Gemini provide
widely used heterogeneous-memory runtime policies~\citep{zhao2023pytorch,fang_you_gemini_colossalai},
and storage-aware systems show that CPU and NVMe can serve as useful capacity
tiers~\citep{lohan,yuan2025cost}. \megatrain treats CPU memory as the authoritative store for parameters and optimizer
states, streams layers through GPU memory, and uses block-wise recomputation to
bound activation residency~\citep{yuan2026megatrain}. However, its activation
schedule remains a fixed heuristic. A fixed checkpoint interval can be feasible,
but it does not search jointly over checkpoint boundaries, GPU/CPU/NVMe homes,
recomputation blocks, and transfer windows.

\input{figs/00_teaser.tex}

The central observation of this paper is that limited-resource training should
be optimized as a constrained scheduling problem. As illustrated in
\figref{fig:motivation}, the bottleneck is not simply that tensors do not fit in
GPU memory. The bottleneck is that GPU compute windows, CPU-GPU traffic,
mandatory parameter and gradient movement, and possible NVMe activation spill
compete for shared resources. A schedule that saves memory by moving tensors to
slower tiers can still hurt throughput if those transfers become visible on the
critical path.

We present \method, an optimization-guided scheduler for limited-resource LLM
training. \method keeps the layer-streaming executor, but
replaces fixed activation scheduling with a mixed-integer model whose decision
variables select checkpoint paths, activation homes, recomputation blocks, and
communication assignments. The objective minimizes recomputation plus the
incremental communication exposure induced by activation placement, rather than
maximizing the amount of offload. The solved policy is loaded once before
training and then executed by the runtime.

Our evaluation covers single-GPU H800 and RTX 3090 experiments from 3B to 27B. Under matched model, data, and batch-size settings, the final \method configuration
reaches 219.95 TFLOPS and 1361 tokens/s, compared with 176.90 TFLOPS and 1075.8
tokens/s for the \megatrain baseline. It peaks at 68.84\,GB of GPU memory and
obtains 95.42\% exact-match accuracy on the full evaluation split. The
\method{} - MILP variant, which removes mixed-integer scheduling while retaining
the other runtime improvements, drops to 193.17 TFLOPS. This is the largest
component-ablation degradation and establishes MILP scheduling as the central
contributor. Removing the Hybrid 8-bit operator, which combines hybrid 8-bit
optimizer states with fast gradient clipping, while retaining MILP scheduling
instead yields 219.29 TFLOPS. The broader experiments report throughput,
memory, maximum feasible batch size, and training quality across model scales
and accelerator budgets.

The contributions of this paper are:
\begin{itemize}
  \item We formulate limited-resource LLM training as a mixed-integer programming(MIP)
  model over checkpoint selection, activation placement, recomputation, and
  CPU-GPU-NVMe communication overlap.
  \item We instantiate this MIP scheduler on top of a 
  layer-streaming executor, which can treat \megatrain baseline as a fixed heuristic schedule
  inside the same scheduling space.
  \item We make a coupled Hybrid 8-bit operator a formal component of \method:
  8-bit optimizer states reduce memory, while fast gradient clipping
  counteracts their additional CPU-side update overhead. We evaluate the
  complete system on H800 and RTX 3090 across model scales.
\end{itemize}

%% file: figs/00_teaser.tex
\begin{figure*}[htbp]
  \centering
  \includegraphics[width=\textwidth]{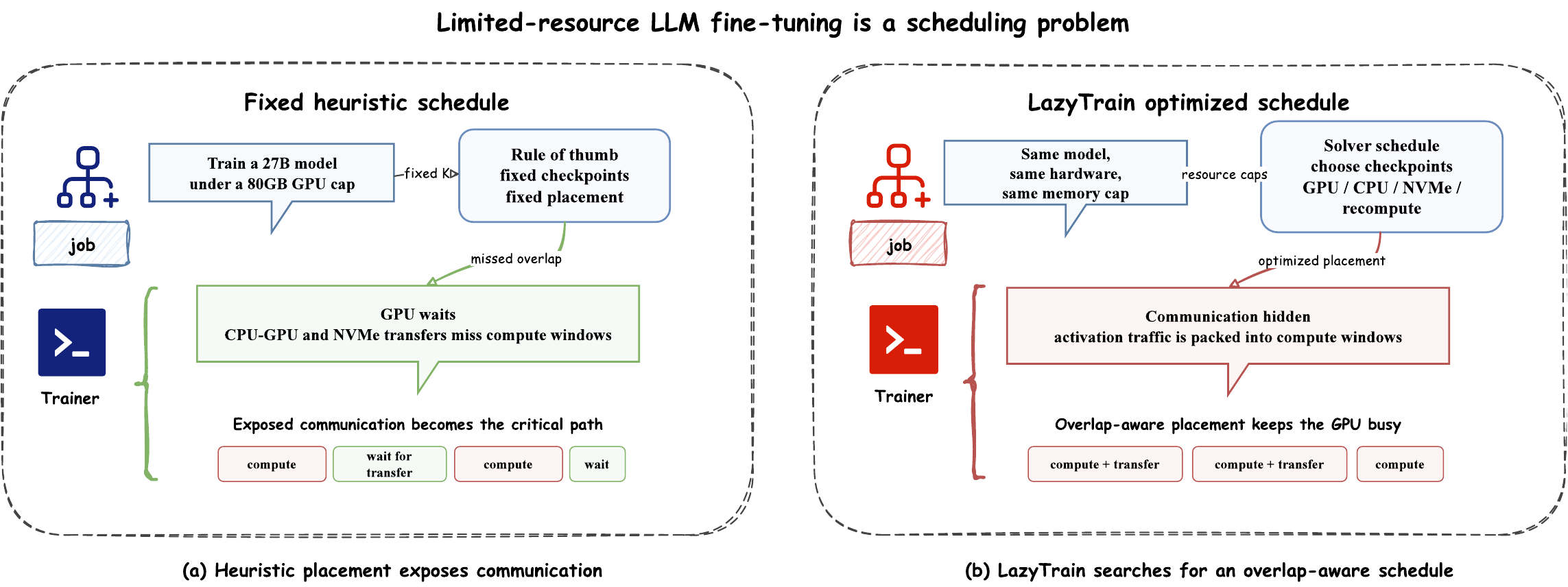}
  \caption{Motivation for optimization-guided training schedules. A fixed
  heuristic can satisfy a memory cap while exposing CPU-GPU or NVMe transfers on
  the critical path. \method instead searches for checkpoint placement and
  transfer assignments that hide activation traffic inside compute windows when
  possible.}
  \label{fig:motivation}
\end{figure*}

%% file: secs/02_related.tex
\section{Related Work}

\parahead{Scaling LLMs and agent workloads}
Recent LLM families continue to push model capacity, context length, and
multimodal capability
\citep{openai2026gpt55,anthropic2026claudeopus47,google2026gemini31pro,deepseekai2026deepseekv4}.
At the same time, agent-oriented models and tools target coding, visual
reasoning, and longer-horizon tool-use workflows
\citep{qwen35blog,team2026kimi,zeng2026glm,anthropic2025claudecode,google2025geminicli,openai2025codexcli}.
Training and adapting these systems increases memory, communication, and
scheduling pressure: larger model states and long-context agent trajectories
raise state and activation footprints, making infrastructure a central part of
model development.

\parahead{Memory-efficient LLM training and offloading}
Efficient LLM training reduces accelerator residency through model parallelism,
mixed precision, recomputation, state sharding, and offloading
\citep{duan2024efficient,shoeybi2019megatron,micikevicius2018mixed,kingma2015adam}.
ZeRO, ZeRO-Infinity, ZeRO++, PyTorch FSDP, and Colossal-AI Gemini partition or
migrate model states across GPU, CPU, and NVMe tiers to make larger models
trainable~\citep{rajbhandari2020zero,rajbhandari2021zero,wang2023zero++,zhao2023pytorch,fang_you_gemini_colossalai}.
Storage-aware systems extend this hierarchy: Ratel optimizes data movement for
fine-tuning large models on consumer GPUs~\citep{lohan}; lifetime-aware
offloading treats GPUDirect Storage as a training tier~\citep{yuan2025cost}; and
heterogeneous tensor caching adds resource-aware migration policies
\citep{afroz202510cache}. \megatrain stores model states in host memory, streams
layers through GPU memory, and uses block-wise recomputation to limit activation
residency~\citep{yuan2026megatrain}.

\parahead{Summary of distinctions}
The systems above make limited-resource training feasible by reducing device
residency, adding heterogeneous memory tiers, or streaming layers through GPU
memory. Among them, \megatrain
chooses activation checkpoints through fixed heuristics. \method addresses a
complementary scheduling question: within the same layer-streaming execution
space, it jointly selects activation checkpoints, tier placement, recomputation,
and exposed communication in one mixed-integer model. It also models NVMe as
both a PCIe consumer and a separate read/write endpoint so that SSD traffic is
used only when it can be hidden or when it enables an otherwise infeasible
schedule.

%% file: secs/03_method.tex
\section{Method}

\subsection{Overview}
\method is an optimization layer over a \megatrain-style layer-streaming
executor. The executor defines the feasible runtime actions: stream parameters
from CPU memory to GPU memory, compute one layer or recomputation block, offload
gradients back to CPU memory, and reload materialized activation checkpoints
from GPU, CPU, or local NVMe. \method decides which of these feasible actions
should be used for each layer boundary before training starts.

\input{figs/01_method_architecture.tex}

\figref{fig:method_architecture} summarizes the method. The scheduler receives
model constants, memory budgets, bandwidth profiles, and the training target.
It solves a mixed-integer linear program (MILP) whose binary variables choose a
checkpoint path and storage homes, while continuous variables assign activation
traffic to compute windows. The solved policy is then loaded by the runtime.
This design keeps execution simple while making the schedule resource-aware.

\subsection{Layer-Boundary Scheduling Space}

Consider a transformer with \(L\) layers and layer-boundary activations
\(\mathcal{B}=\{0,\ldots,L\}\). Boundary 0 is the block input, and boundary
\(i\), for \(1\le i\le L\), is the output of layer \(i\). Let \(A_i\) be the
activation size at boundary \(i\), and let \(\rho_j\) be the cost of the forward
transition from boundary \(j\) to boundary \(j+1\). A directed edge
\((s,e)\), where \(0 \le s < e \le L\), represents one backward recomputation
block: boundaries \(s\) and \(e\) are materialized, while the layers between
them can be recomputed locally. The set of candidate edges is
\(\mathcal{E}\).

For each edge, \method computes recomputation cost as
\begin{equation}
R_{s,e} =
\begin{cases}
0, & e \le s+1,\\
\sum_{j=s}^{e-2}\rho_j, & e > s+1.
\end{cases}
\label{eq:lazy_recompute_cost}
\end{equation}
The final layer output of the block is already available at boundary \(e\), so
the cost excludes that output. This edge view turns activation checkpointing
into a path-selection problem.

\subsection{Decision Variables and Constraints}

\method uses binary variables \(x_{s,e}\) to select recomputation edges and
\(z_i\) to indicate whether boundary \(i\) is materialized. If a boundary is
materialized, binary variables \(g_i\), \(c_i\), and \(n_i\) assign it to GPU
high-bandwidth memory (HBM), CPU DRAM, or local NVMe:
\begin{equation}
z_i = g_i + c_i + n_i,\qquad i\in\mathcal{B}.
\label{eq:lazy_unique_home}
\end{equation}
The executor keeps the input and output boundaries in HBM, so
\(g_0=g_L=1\) and \(c_0=c_L=n_0=n_L=0\).
The selected edges must form one legal path from the input boundary to the
output boundary:
\begin{equation}
\begin{aligned}
\sum_{(0,e)\in\mathcal{E}} x_{0,e} &= 1,&
\sum_{(s,L)\in\mathcal{E}} x_{s,L} &= 1,\\
\sum_{(s,i)\in\mathcal{E}} x_{s,i} &= z_i,&
\sum_{(i,e)\in\mathcal{E}} x_{i,e} &= z_i,\;0<i<L.
\end{aligned}
\label{eq:lazy_path}
\end{equation}
Placement must also satisfy tier capacities:
\begin{equation}
\begin{aligned}
\sum_i A_i g_i &\le M_G,\\
\sum_i A_i c_i &\le M_C,\\
\sum_i A_i n_i &\le M_N,
\end{aligned}
\label{eq:lazy_memory}
\end{equation}
where \(M_G\), \(M_C\), and \(M_N\) are activation budgets for GPU, CPU, and
NVMe. In the H800 27B run, the GPU activation budget is derived from a 70\,GB
peak-memory cap.

\subsection{Communication Resource Model}

Activation offload and reload share the same PCIe fabric as mandatory parameter
prefetch and gradient offload. \method therefore assigns activation transfer to
compute windows \(w\in\{0,\ldots,L-1\}\) and penalizes only the traffic that
cannot be hidden. Let
\(d_{i,w}\) and \(h_{i,w}\) be activation device-to-host (D2H) and
host-to-device (H2D) traffic for boundary \(i\) assigned to window \(w\). Let
\(G_w\) and \(P_w\) be mandatory gradient
D2H and parameter H2D traffic, and let \(C^D_w\) and \(C^H_w\) be the D2H and
H2D capacity hidden by compute in window \(w\). The PCIe constraints are
\begin{equation}
\begin{aligned}
\sum_{w=i}^{L-1} d_{i,w} &= A_i(c_i+n_i),\\
\sum_{w=i}^{L-1} h_{i,w} &= A_i(c_i+n_i),\\
\sum_{i\le w} d_{i,w}+G_w &\le C^D_w+\epsilon^D_w,\\
\sum_{i\le w} h_{i,w}+P_w &\le C^H_w+\epsilon^H_w.
\end{aligned}
\label{eq:lazy_pcie}
\end{equation}
Mandatory traffic can exceed a window's overlap capacity even without
activation offloading. Define constant baseline slacks
\(\bar{\epsilon}^{D}_w=\max\{0,G_w-C^D_w\}\) and
\(\bar{\epsilon}^{H}_w=\max\{0,P_w-C^H_w\}\), and let
\(\Delta^D_w=\epsilon^D_w-\bar{\epsilon}^D_w\) and
\(\Delta^H_w=\epsilon^H_w-\bar{\epsilon}^H_w\) denote the additional exposure
caused by activation placement.
NVMe placements additionally consume SSD endpoint bandwidth. With NVMe write
traffic \(q_{i,w}\), read traffic \(r_{i,w}\), and endpoint capacities
\(C^W_w,C^R_w\), \method enforces
\begin{equation}
\begin{aligned}
\sum_{w=i}^{L-1} q_{i,w} &= A_i n_i,\\
\sum_{w=i}^{L-1} r_{i,w} &= A_i n_i,\\
\sum_{i\le w} q_{i,w} &\le C^W_w+\epsilon^W_w,\\
\sum_{i\le w} r_{i,w} &\le C^R_w+\epsilon^R_w.
\end{aligned}
\label{eq:lazy_nvme}
\end{equation}
This separates the PCIe fabric from the NVMe endpoint. A GPUDirect Storage
(GDS)-style NVMe reload is useful only if both resources can accommodate it
without exposing slow I/O.

\subsection{Objective and Execution}

The scheduler minimizes recomputation plus incremental exposed communication:
\begin{equation}
\begin{aligned}
\min\quad
&\sum_{(s,e)\in\mathcal{E}} R_{s,e}x_{s,e}
+ \lambda_z\sum_i z_i
+ \lambda_n\sum_i n_i\\
&+\sum_w\left(
\frac{\Delta^D_w}{B_D}
+\frac{\Delta^H_w}{B_H}
+\frac{\epsilon^W_w}{B_W}
+\frac{\epsilon^R_w}{B_R}
\right),
\end{aligned}
\label{eq:lazy_objective}
\end{equation}
where \(B_D,B_H,B_W,B_R\) convert bytes into time-like costs. The PCIe baseline
subtraction charges only activation-induced exposure; \(\Delta^{D/H}_w\) is
nonnegative by Eq.~\eqref{eq:lazy_pcie}. No baseline is needed for the NVMe
terms because Eq.~\eqref{eq:lazy_nvme} contains only activation traffic. The
small nonnegative \(\lambda_z\) and \(\lambda_n\) are tie-breakers that avoid
unnecessary materialized and NVMe checkpoints. Thus, the objective minimizes
activation-induced stalls while accounting for recomputation, rather than
maximizing offload volume.
Algorithm~\ref{alg:lazy_milp_build} summarizes the complete construction from
model parsing and variable creation to SCIP solving and schedule generation.

\begin{algorithm}[htbp]
\caption{\method MILP Construction}
\label{alg:lazy_milp_build}
\begin{algorithmic}[1]
\Require Model configuration, batch size \(B\), sequence length \(S\), budgets
\(M_G,M_C,M_N\), bandwidth windows, recomputation costs \(\rho\)
\Ensure Solved schedule or infeasibility status
\State Parse the model configuration to obtain \(L\), hidden size, bytes per
element, and layer parameter counts.
\State Compute activation sizes \(A_i\) and static memory terms for parameters,
gradients, optimizer states, and runtime reserve.
\State Build candidate boundary set \(\mathcal{B}\) and segment set
\(\mathcal{E}\) subject to the maximum recomputation length.
\State Create binary variables \(x_{s,e}\), \(z_i\), \(g_i\), \(c_i\), and
\(n_i\) for all legal segments, boundaries, and tiers.
\State Create continuous traffic variables for CPU D2H/H2D and NVMe write/read
assignments, plus exposed-communication slack variables.
\State Add path-flow constraints so selected segments form one backward path
from boundary \(0\) to boundary \(L\).
\State Add unique-placement constraints \(z_i=g_i+c_i+n_i\) for every boundary
\(i\).
\State Add HBM, CPU DRAM, and NVMe capacity constraints.
\State Add PCIe and NVMe endpoint bandwidth-window constraints, including
mandatory parameter H2D and gradient D2H traffic.
\State Minimize recomputation cost plus activation-induced communication
exposure, with small tie-breaking penalties for otherwise equivalent schedules.
\State Call SCIP through PySCIPOpt and retain the incumbent schedule if the
solver status is feasible or optimal.
\end{algorithmic}
\end{algorithm}

After solving the mixed-integer problem once before training, the runtime
follows the selected policy during forward and backward execution. It keeps
parameters and optimizer states in CPU memory, uses GPU memory as a transient
compute cache, reloads selected activation checkpoints from their assigned
tiers, and recomputes omitted blocks as specified by the path. The Hybrid 8-bit
operator is the complete optimizer-side component: it stores a small subset of
optimizer states in CPU 8-bit AdamW format and performs fast gradient clipping.
The two mechanisms are deliberately coupled. On our single-GPU CPU-offload
path, 8-bit optimizer states primarily reduce CPU-resident optimizer memory;
they do not by themselves guarantee faster steps because quantization and
dequantization, block-wise scale handling, and 8-bit state updates add CPU
work. Fast gradient clipping reduces clipping overhead to counteract this added
optimizer-side cost. We therefore define and ablate the two mechanisms as one
formal \method component rather than as independent external optimizations.
In the measured full-system configuration, DeepSpeed CPUAdam handles most
parameters, while a 2\% parameter slice uses block-wise 8-bit AdamW states with
block size 4096. The clipping path computes per-tensor norms with CPU foreach
kernels, aggregates the global norm in FP32, and applies the scale with a
foreach multiplication; unsupported or non-finite cases fall back to
per-tensor FP32 accumulation.
Table~\ref{tab:lazy_nvme_schedule} reports the final schedule used by the
measured Qwen3.6-27B training run. For this configuration, the solver fixes the
recomputation edges and GPU-checkpoint set from the \method{} - Hybrid 8-bit
variant, then optimizes the remaining CPU/NVMe homes and transfer-window
assignments. The reported solver status therefore certifies optimality within
this restricted subproblem, not over the full MILP space defined above.

\begin{table}[htbp]
\centering
\small
\setlength{\tabcolsep}{3.5pt}
\caption{Final restricted NVMe-aware schedule for Qwen3.6-27B supervised
fine-tuning with batch size 72 and sequence length 1024. Values are the schedule
constants used by the reported training run.}
\label{tab:lazy_nvme_schedule}
\resizebox{\linewidth}{!}{
\begin{tabular}{lcl}
\toprule
\textbf{Item} & \textbf{Value} & \textbf{Meaning} \\
\midrule
\rowcolor{gray!30}
\multicolumn{3}{l}{\textbf{Model and resource budget}} \\
Layers & 64 & Transformer blocks \\
Activation / boundary & 0.703\,GiB & One hidden-state checkpoint \\
GPU activation budget & 21.55\,GB & From 70\,GB peak cap \\
CPU / NVMe budget & 15 / 32\,GB & Host and SSD activation caps \\
\rowcolor{gray!30}
\multicolumn{3}{l}{\textbf{Schedule solution}} \\
Checkpoints & 52 & Materialized boundaries \\
GPU / CPU / NVMe & 30 / 21 / 1 & Tier placement counts \\
Solve scope & Restricted & Recomputation edges and GPU set fixed \\
NVMe boundary & 21 & Checkpoint placed on SSD \\
\rowcolor{gray!30}
\multicolumn{3}{l}{\textbf{Transfer rates and exposure}} \\
PCIe H2D/D2H rate & 12 / 12\,GB/s & Measured rate used by scheduler \\
NVMe read/write rate & 2.83 / 3.35\,GB/s & Measured endpoint rate \\
Exposed PCIe/NVMe comm. & 0.0 / 0.0\,ms & Incremental exposed traffic \\
\bottomrule
\end{tabular}}
\end{table}

%% file: figs/01_method_architecture.tex
\begin{figure*}[htbp]
  \centering
  \includegraphics[width=\textwidth]{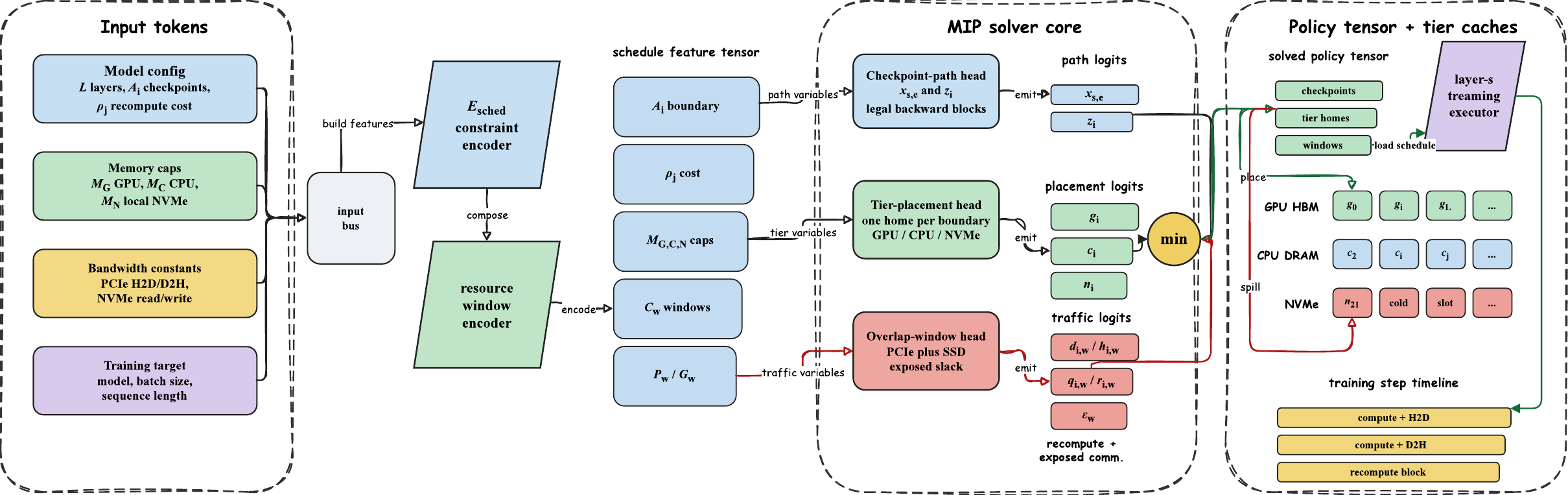}
  \caption{\method architecture. Resource profiles, model configuration,
  and training targets define a mixed-integer scheduling problem. The solver
  chooses checkpoint boundaries, GPU/CPU/NVMe homes, and transfer windows; the
  resulting policy is executed by a layer-streaming runtime.}
  \label{fig:method_architecture}
\end{figure*}

%% file: secs/04_experiments.tex
\section{Experiments}

\subsection{Experimental Setup}

We evaluate \method through single-GPU supervised fine-tuning experiments on
H800 80GB and RTX 3090 24GB accelerators. The primary matched comparison uses
Qwen3.6-27B, a checkpoint with a Qwen3.5-style text
architecture~\citep{qwen35blog}, and MetaMathQA with a 70/30
training/evaluation split, sequence length 1024, one training epoch, batch size
72, and 3841 optimizer steps. The compared 27B variants share the same hardware,
model, data split, sequence length, and batch size. The broader H800 and RTX
3090 results are also obtained from experiments under their corresponding
model and hardware settings. Unless stated otherwise, each row reports one run.

\begin{table}[htbp]
\centering
\small
\setlength{\tabcolsep}{3.5pt}
\caption{H800 node and transfer rates used in the \method experiments.}
\label{tab:h800_environment}
\resizebox{\linewidth}{!}{
\begin{tabular}{lcc}
\toprule
\textbf{Tier / component} & \textbf{Capacity / topology} & \textbf{Rate / scope} \\
\midrule
GPU HBM & 80\,GB & Single-GPU runs \\
Host DRAM & 2.0\,TiB & 128 logical CPUs \\
CPU--GPU link & PCIe Gen5 x16 & 12\,GB/s per direction \\
Local NVMe SSDs & 2\(\times\)7.0\,TB & 2.83 / 3.35\,GB/s read/write \\
\bottomrule
\end{tabular}}
\end{table}

Table~\ref{tab:h800_environment} summarizes the node capacity and topology
together with the transfer rates used by the scheduler. We report maximum
feasible batch size, sustained TFLOPS, GPU memory, CPU memory, and tokens per
second. Accuracy is reported separately because throughput and exact-match
accuracy answer different questions. For the H800 runs, sustained
TFLOPS follows the \megatrain accounting convention and is computed from
complete-batch training steps. Tokens/s is reported as a run-level throughput
value.

\subsection{Training Throughput}

\input{figs/02_training_performance.tex}

Panels (a) and (b) of \figref{fig:training_performance}, together with
\tabref{tab:h800_training_perf}, report the H800 experiments from 3B to 27B. In
the primary matched 27B pair, \method improves sustained throughput from 176.90
to 219.95 TFLOPS, a 1.24$\times$ increase, while using the same model, data
split, sequence length, and batch size. The 3B--14B experiments show the same
direction of improvement over \megatrain, with \method reaching 176.42--212.34
TFLOPS across these model scales.

Panels (c) and (d) report the constrained RTX 3090 experiments. At every model
scale, \method increases the maximum feasible batch size by one over
\megatrain and yields higher token throughput. ZeRO-3 Offload runs out of memory
at batch size 1 for the 14B and 27B models. These results provide empirical
evidence that the scheduling benefit persists under a substantially smaller
GPU memory budget.

\begin{table}[htbp]
\centering
\scriptsize
\setlength{\tabcolsep}{1.5pt}
\caption{Experimental training performance on H800 and RTX 3090. Models
3B--14B are Qwen2.5; 27B is Qwen3.6. Z3 and Mega abbreviate ZeRO-3 Offload and
\megatrain. The H800 27B Mega/\method pair
comprises matched runs. OOM denotes an observed out-of-memory result at batch
size 1. GPU and CPU memory are reported in GB.}
\label{tab:h800_training_perf}
\resizebox{\linewidth}{!}{%
\begin{tabular}{llccccc}
\toprule
\textbf{Model} & \textbf{System} & \textbf{Batch} & \textbf{TFLOPS} &
\textbf{GPU} & \textbf{CPU} & \textbf{tok/s} \\
\midrule
\rowcolor{gray!30}
\multicolumn{7}{l}{\textbf{H800}} \\
3B & Z3 & 32 & 68.65 & 74.12 & 61.8 & 3814 \\
3B & Mega & 192 & 142.80 & 35.72 & 49.31 & 7933 \\
3B & \method & 192 & \textbf{176.42} & 67.95 & 52.68 & \textbf{9801} \\
\addlinespace[2pt]
7B & Z3 & 16 & 79.88 & 76.43 & 124.6 & 1902 \\
7B & Mega & 144 & 161.35 & 43.26 & 95.74 & 3842 \\
7B & \method & 144 & \textbf{199.84} & 68.21 & 99.58 & \textbf{4758} \\
\addlinespace[2pt]
14B & Z3 & 8 & 93.43 & 77.21 & 257.9 & 1112 \\
14B & Mega & 96 & 171.92 & 52.83 & 179.64 & 2047 \\
14B & \method & 96 & \textbf{212.34} & 68.57 & 185.12 & \textbf{2528} \\
\addlinespace[2pt]
27B & Z3 & 2 & 86.55 & 78.34 & 518.6 & 534 \\
27B & Mega & 72 & 176.90 & 60.40 & 339.99 & 1075.8 \\
27B & \method & 72 & \textbf{219.95} & 68.84 & 361.58 & \textbf{1361} \\
\midrule
\rowcolor{gray!30}
\multicolumn{7}{l}{\textbf{RTX 3090}} \\
3B & Z3 & 1 & 23.91 & 20.32 & -- & -- \\
3B & Mega & 7 & 33.18 & 22.83 & 25.0 & 1792 \\
3B & \method & 8 & \textbf{40.76} & 23.16 & 27.8 & \textbf{2210} \\
\addlinespace[2pt]
7B & Z3 & 1 & 27.49 & 20.83 & -- & -- \\
7B & Mega & 5 & 35.09 & 22.63 & 56.7 & 768 \\
7B & \method & 6 & \textbf{43.42} & 23.06 & 61.5 & \textbf{951} \\
\addlinespace[2pt]
14B & Z3 & 1 & \textbf{OOM} & -- & -- & -- \\
14B & Mega & 3 & 30.19 & 21.10 & 103.7 & 341 \\
14B & \method & 4 & \textbf{37.36} & 22.92 & 111.8 & \textbf{422} \\
\addlinespace[2pt]
27B & Z3 & 1 & \textbf{OOM} & -- & -- & -- \\
27B & Mega & 1 & 22.74 & 22.48 & 197.3 & 140 \\
27B & \method & 2 & \textbf{28.11} & 23.08 & 214.6 & \textbf{174} \\
\bottomrule
\end{tabular}}
\end{table}

\begin{table}[htbp]
\centering
\small
\setlength{\tabcolsep}{2.4pt}
\caption{Experimental exact-match accuracy comparison. Every cell is an
evaluation result from the corresponding model and system experiment; the 27B
\method cell uses the full evaluation split. Higher is better.}
\label{tab:accuracy_compare}
\resizebox{\linewidth}{!}{%
\begin{tabular}{lcccc}
\toprule
\textbf{Metric} & \textbf{ZeRO-3 Offload} & \textbf{ZeRO-Infinity} &
\textbf{\megatrain} & \textbf{\method} \\
\midrule
7B Acc. (\%) & 88.93 & 88.97 & \textbf{88.99} & 88.95 \\
14B Acc. (\%) & 92.41 & 92.36 & \textbf{92.52} & 92.47 \\
27B Acc. (\%) & 95.27 & 95.31 & 95.33 & \textbf{95.42} \\
\bottomrule
\end{tabular}}
\end{table}

\subsection{Schedule Case Study}

\input{figs/06_qwen_json_schedule.tex}

Figure~\ref{fig:qwen_json_schedule_map} shows a separate Qwen3.6-27B schedule
found with an 8\,GB CPU activation budget and a 32\,GB NVMe budget. SCIP
returned this valid incumbent at the time limit: 30 checkpoints remain in HBM,
11 move to CPU DRAM, 11 move to NVMe, and 13 boundaries are recomputed. This
schedule experiment is separate from the final schedule in
Table~\ref{tab:lazy_nvme_schedule}, which uses a 15\,GB CPU activation budget
and places only boundary 21 on NVMe. The case illustrates how the experimental
placement changes under a tighter CPU budget. Additional solver cases are
provided in Appendix~\ref{sec:app_solver_case_study}.

\subsection{Component Ablation}

\input{figs/04_component_ablation.tex}

\figref{fig:component_ablation} and \tabref{tab:component_ablation} isolate the
main 27B variants, where a minus sign denotes removal of the named component.
The complete \method configuration reaches 219.95 TFLOPS and 1361 tokens/s.
Removing MILP-based scheduling while retaining the other runtime improvements
drops performance to 193.17 TFLOPS and 1195 tokens/s, reductions of 12.2\% in
both metrics. This is the largest component-ablation degradation, showing that
joint checkpoint, placement, recomputation, and communication scheduling is the
primary source of the gain. Removing the complete Hybrid 8-bit operator---both
the hybrid 8-bit optimizer states and fast gradient clipping---while retaining
MILP scheduling yields 219.29 TFLOPS and 1357 tokens/s, 0.3\% below the complete
system. We remove these mechanisms together because they form one operator:
8-bit states save optimizer memory but add CPU update work, whereas fast
gradient clipping counteracts the associated step overhead. Thus, Hybrid 8-bit
is a formal part of \method, while the MILP scheduler remains its primary method
innovation and most important component in this configuration. The \method{} -
MILP run reports throughput and memory.

\begin{table}[htbp]
\centering
\small
\setlength{\tabcolsep}{3.5pt}
\caption{Qwen3.6-27B component ablation on H800. Rows are reported runs under
the same model, data split, sequence length, and batch size. A minus sign denotes removal of the named component from the
complete \method system.}
\label{tab:component_ablation}
\resizebox{\linewidth}{!}{%
\begin{tabular}{lccccc}
\toprule
\textbf{Method} & \textbf{Max BS} & \textbf{TFLOPS} &
\textbf{GPU Mem} & \textbf{CPU Mem} & \textbf{tok/s} \\
\midrule
\megatrain baseline & 72 & 176.90 & 60.40\,GB & 339.99\,GB & 1075.8 \\
\method & 72 & \textbf{219.95} & 68.84\,GB & 361.58\,GB & \textbf{1361} \\
\method{} - MILP & 72 & 193.17 & 68.84\,GB & 362.01\,GB & 1195 \\
\method{} - Hybrid 8-bit & 72 & 219.29 & 68.84\,GB & 360.23\,GB & 1357 \\
\bottomrule
\end{tabular}}
\end{table}

\subsection{Training Quality}

\input{figs/05_accuracy_quality.tex}

\figref{fig:accuracy_quality} and Table~\ref{tab:accuracy_compare} report
experimental exact-match results for all compared systems. The 27B \method
score, 95.42\%, uses the full MetaMathQA evaluation split.

\subsection{Discussion}

Across H800 and RTX 3090, solver-selected activation scheduling improves
throughput over the fixed \megatrain baseline. In the 27B ablation, the
largest degradation occurs when MILP scheduling is removed: throughput falls
by 12.2\% from 219.95 to 193.17 TFLOPS. Removing the Hybrid 8-bit operator
produces a smaller 0.3\% drop. The final NVMe-aware schedule expands the feasible
placement space with zero solver-reported incremental communication exposure.

%% file: figs/02_training_performance.tex
\begin{figure*}[htbp]
  \centering
  \includegraphics[width=\textwidth]{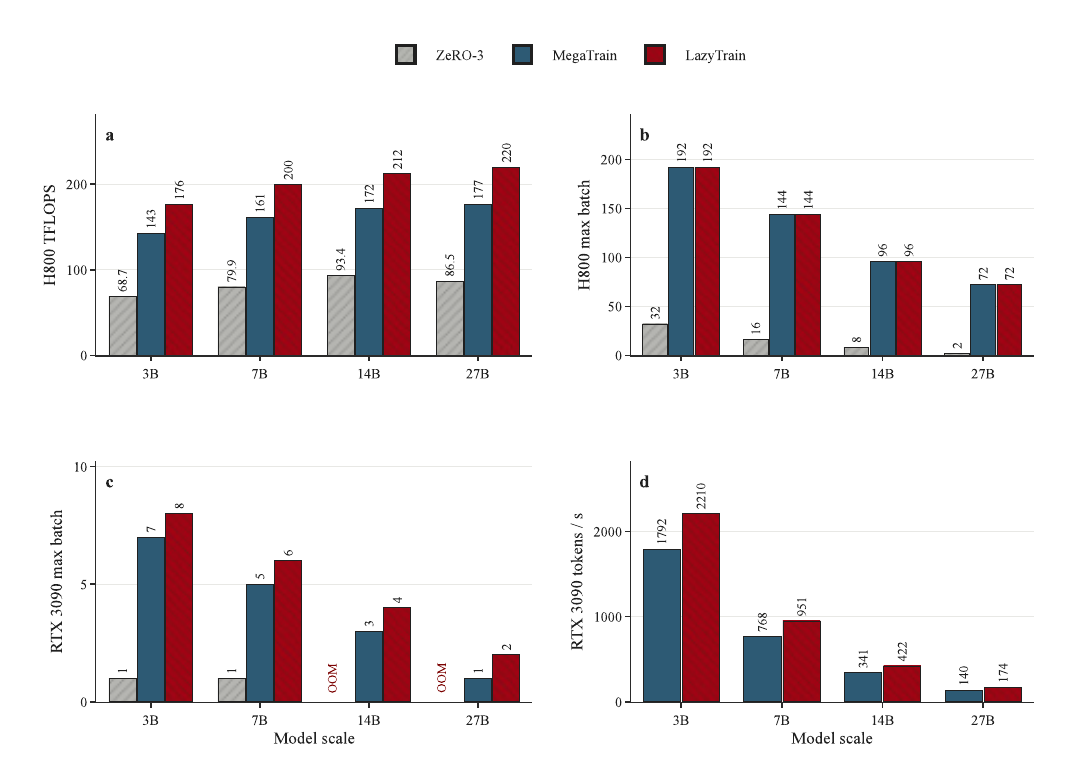}
  \caption{Experimental training performance on H800 and RTX 3090 across model
  scales.The H800 27B \megatrain/\method pair is the primary matched
  comparison. OOM denotes an observed out-of-memory result at batch size 1.}
  \label{fig:training_performance}
\end{figure*}

%% file: figs/06_qwen_json_schedule.tex
\begin{figure*}[htbp]
\centering
\small
\begin{tcolorbox}[colback=gray!4!white, colframe=blue!65!black,
title=LazyTrain Solver Case: Qwen3.6-27B boundary-wise schedule,
boxrule=0.3mm, width=\textwidth, arc=3mm, auto outer arc=true]
\centering
\begin{tikzpicture}[
  x=0.52cm,
  y=0.52cm,
  tile/.style={draw, rounded corners=0.45mm, minimum width=0.46cm,
    minimum height=0.42cm, inner sep=0pt, font=\tiny},
  note/.style={draw, rounded corners=1.1mm, align=left, inner sep=3pt,
    font=\scriptsize, text width=0.325\textwidth}
]
\tikzset{
  G/.style={fill=blue!18, draw=blue!70!black},
  C/.style={fill=orange!18, draw=orange!80!black},
  N/.style={fill=red!16, draw=red!70!black},
  R/.style={fill=gray!22, draw=gray!70!black}
}
\node[anchor=west, font=\bfseries\scriptsize] at (0,1.15)
  {Activation-boundary placement map};
\node[note, fill=blue!5, draw=blue!60!black, anchor=north west] at (14.3,1.20)
  {\textbf{Input and solver}\\
  Qwen3.6-27B; \(L=64\), \(B=72\), \(S=1024\).\\
  CPU/NVMe budgets: 8/32\,GB.\\
  SCIP: \texttt{timelimit}; objective \(182.00052\).};
\node[note, fill=green!6, draw=green!45!black, anchor=north west] at (14.3,-1.35)
  {\textbf{Materialized plan}\\
  52 stored checkpoints = 30 GPU + 11 CPU + 11 NVMe.\\
  13 boundaries are recomputed.};
\node[note, fill=red!5, draw=red!65!black, anchor=north west] at (14.3,-3.65)
  {\textbf{Budget and overlap}\\
  Peak cap/solver-reported peak: 70/70\,GB.\\
  Activation budget: 21.55\,GB.\\
  Exposed communication: 0\,ms.};
\foreach \row/\range in {0/{0--12},1/{13--25},2/{26--38},3/{39--51},4/{52--64}} {
  \node[anchor=east, font=\tiny] at (-0.35,-\row) {\range};
}
\foreach \i/\s/\row/\col in {
0/G/0/0,1/R/0/1,2/N/0/2,3/R/0/3,4/N/0/4,5/G/0/5,6/C/0/6,
7/C/0/7,8/G/0/8,9/C/0/9,10/G/0/10,11/C/0/11,12/C/0/12,
13/R/1/0,14/G/1/1,15/N/1/2,16/C/1/3,17/C/1/4,18/G/1/5,
19/C/1/6,20/C/1/7,21/G/1/8,22/G/1/9,23/G/1/10,24/N/1/11,25/G/1/12,
26/R/2/0,27/N/2/1,28/G/2/2,29/R/2/3,30/C/2/4,31/C/2/5,
32/G/2/6,33/G/2/7,34/G/2/8,35/N/2/9,36/G/2/10,37/R/2/11,38/G/2/12,
39/N/3/0,40/N/3/1,41/R/3/2,42/R/3/3,43/R/3/4,44/N/3/5,
45/G/3/6,46/G/3/7,47/G/3/8,48/G/3/9,49/G/3/10,50/R/3/11,51/N/3/12,
52/G/4/0,53/G/4/1,54/G/4/2,55/G/4/3,56/N/4/4,57/R/4/5,
58/R/4/6,59/R/4/7,60/G/4/8,61/G/4/9,62/G/4/10,63/G/4/11,64/G/4/12} {
  \node[tile,\s] at (\col,-\row) {\i};
}
\node[tile,G,label=right:{\scriptsize GPU}] at (0,-5.55) {};
\node[tile,C,label=right:{\scriptsize CPU}] at (3.0,-5.55) {};
\node[tile,N,label=right:{\scriptsize NVMe}] at (6.0,-5.55) {};
\node[tile,R,label=right:{\scriptsize Recompute}] at (9.3,-5.55) {};
\end{tikzpicture}
\end{tcolorbox}
\caption{Boundary-wise JSON schedule map for Qwen3.6-27B on MetaMathQA at batch
size 72. This time-limit incumbent is a separate schedule experiment using
8/32\,GB CPU/NVMe activation budgets rather than the final schedule in
Table~\ref{tab:lazy_nvme_schedule}.
Blue tiles remain in GPU HBM, amber tiles move to CPU DRAM, red tiles move to
NVMe, and gray tiles are recomputed.}
\label{fig:qwen_json_schedule_map}
\end{figure*}
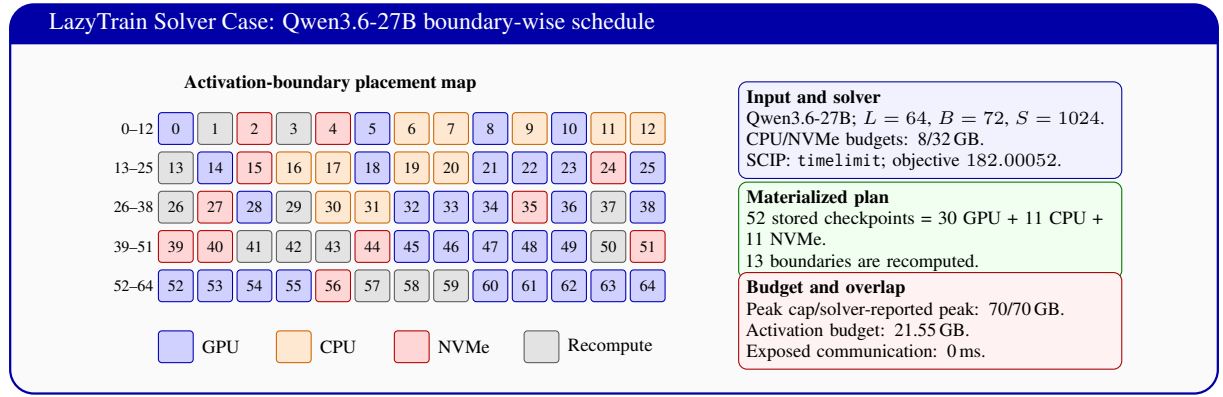

%% file: figs/04_component_ablation.tex
\begin{figure}[htbp]
  \centering
  \includegraphics[width=\linewidth]{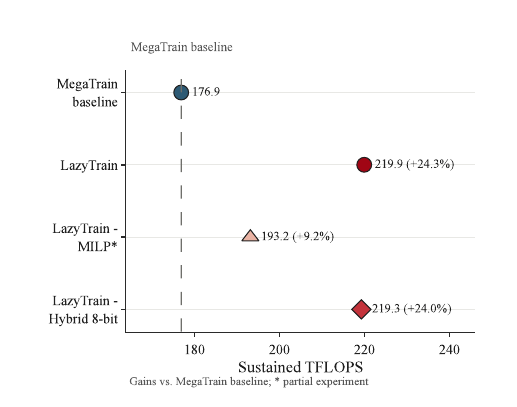}
  \caption{Qwen3.6-27B component ablation on H800. Markers show runs at
  batch size 72, and percentages are throughput gains over the dashed
  \megatrain baseline. A minus sign denotes removal of the named \method
  component.}
  \label{fig:component_ablation}
\end{figure}

%% file: figs/05_accuracy_quality.tex
\begin{figure}[htbp]
  \centering
  \includegraphics[width=\linewidth]{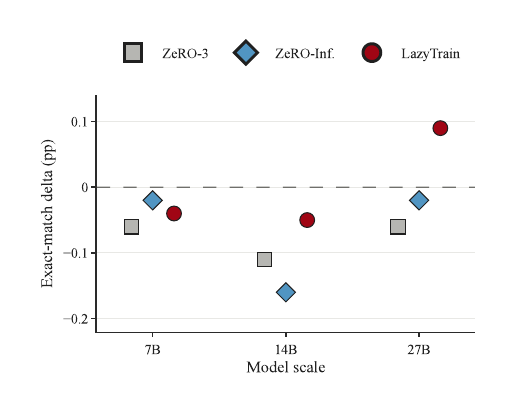}
  \caption{Experimental exact-match accuracy deltas relative to \megatrain
  across 7B, 14B, and 27B models. All plotted values are evaluation results; the
  27B \method value is measured on the full evaluation split. The dashed line
  denotes the \megatrain reference.}
  \label{fig:accuracy_quality}
\end{figure}

%% file: secs/09_conclusion.tex
\section{Conclusion}

\method centers on a mixed-integer scheduler, its primary method innovation,
that jointly optimizes activation checkpointing, tier placement, recomputation,
and CPU-GPU-NVMe communication overlap under limited hardware resources.
It improves sustained TFLOPS over
\megatrain by approximately 1.24$\times$ across H800 models; on RTX 3090, it
achieves higher throughput and a one-unit larger feasible batch size. The
primary matched 27B H800 run reaches 219.95 TFLOPS under a 70\,GB peak cap.
The component ablation further identifies MILP scheduling as the dominant
component: removing it reduces throughput by 12.2\%, from 219.95 to 193.17
TFLOPS. The complete framework also includes the Hybrid 8-bit operator, which
couples memory-saving 8-bit optimizer states with fast gradient clipping to
counteract their added CPU-side update overhead; removing the operator reduces
throughput to 219.29 TFLOPS.

\clearpage
\section*{Limitations}
The experiments are limited to single-GPU H800 and RTX 3090 settings. Individual
The 30\% evaluation split is also used for
periodic loss monitoring, so the final exact-match value is not evaluated on an
untouched held-out test set. Exposed-communication values are recorded solver
outputs under the experimental resource and bandwidth settings; the runtime
evaluation does not separately instrument per-step stall time. The scheduler is
offline and does not adapt to runtime bandwidth variation. Multi-GPU,
multi-node, and online scheduling remain future work.

\section*{Ethical Considerations}
This work studies systems mechanisms for training large language models under
limited hardware resources. The experiments use a mathematical-reasoning
benchmark and system measurements; we collect no new human-subject data.
Improving training efficiency can broaden access to large-model research and
reduce wasted computation, but lowering the hardware barrier can also make it
easier to train models for harmful uses. \method makes its placement and
recomputation decisions explicit through solver schedules, but it does not
mitigate risks inherited from the base model, training data, or downstream
application. Deployments should therefore retain task-appropriate data
governance, access control, logging, safety evaluation, and human oversight.

\section*{Information About Use of AI Assistants}
In preparing this manuscript, the authors used AI-assisted technology,
specifically large language models such as GPT-5 and DeepSeek-V4, exclusively
for text refinement. The tools assisted with proofreading, grammatical
correction, and polishing linguistic expressions to improve the clarity and
readability of the manuscript. The authors are responsible for the final
content, claims, and verification.

%% file: appx.tex
\appendix

\section*{Appendix Overview}
This appendix provides supporting details for the method, experiments, and
solver analyses in the main paper. It is organized as follows. 
\begin{itemize}
\item Appendix~\ref{sec:app_solver_cases}
reports representative solver outputs and placement decisions. 
\item Appendix~\ref{sec:app_milp_solution}
describes the canonical MILP and its branch-and-cut solution procedure.
\item Appendix~\ref{sec:app_experimental_configuration} specifies the resource
model, training protocol, and scheduler variants. 
\item Appendix~\ref{sec:app_server_environment}
records the server hardware and software environment. 
\item Appendix~\ref{sec:app_reproducibility}
collects the training protocol, hardware checks, and reproducibility package.
\item Appendix~\ref{sec:app_broader_impact} discusses broader impact and compute
resources. 
\item Appendix~\ref{sec:app_solver_case_study} presents visual
Qwen3.6-27B and GPT-OSS-120B case studies that make placement choices,
feasibility limits, and communication bottlenecks concrete.
\end{itemize}

\input{secs/10_solver_appendix.tex}
\input{secs/11_reproducibility.tex}
\input{secs/12_impact.tex}
\input{secs/13_case_study.tex}

%% file: secs/10_solver_appendix.tex
\section{Solver Outputs and Placement Details}
\label{sec:app_solver_cases}

This appendix gives a concrete view of what the \method solver returns before
training starts. The Qwen3.6-27B row is the final NVMe-aware schedule used for
the H800 MetaMathQA experiment. This final run fixes the \method{} - Hybrid 8-bit
recomputation edges and GPU-checkpoint set, then solves the remaining CPU/NVMe
homes and transfer assignments. The GPT-OSS-120B mixture-of-experts (MoE) rows
are admission-control and placement experiments instantiated from the Hugging
Face model configuration. They report the observed SCIP outcomes under the
specified CPU-memory and optimizer-state settings.

\begin{table*}[t]
\centering
\scriptsize
\setlength{\tabcolsep}{3pt}
\caption{Representative solver-experiment outcomes. ``Mandatory exposed'' is
the solver-reported parameter/gradient traffic that cannot be hidden by the
configured compute windows. It is reported for diagnosis but subtracted as a
constant baseline from the placement objective.}
\label{tab:appendix_solver_snapshots}
\begin{tabularx}{\textwidth}{@{}p{0.17\textwidth}p{0.10\textwidth}p{0.17\textwidth}XX@{}}
\toprule
\textbf{Model / setting} & \textbf{Status} & \textbf{Budget} &
\textbf{Solver result} & \textbf{Interpretation} \\
\midrule
\rowcolor{gray!30}
\multicolumn{5}{l}{\textbf{Qwen3.6-27B schedule}} \\
Qwen3.6-27B dense; H800 MetaMathQA; batch 72 &
Optimal within restricted solve &
Peak $\le 70$GB; CPU activation 15GB; NVMe activation 32GB &
23.68B trainable parameters; activation 0.703GiB per checkpoint; 52
checkpoints: 30 GPU, 21 CPU, 1 NVMe, 13 recompute boundaries; objective
182.0015; exposed communication: activation/NVMe/mandatory = 0/0/0ms &
With recomputation edges and GPU checkpoints fixed, the final 27B schedule fits the
70GB HBM cap; the solver reports all activation movement hidden inside compute
windows. \\
\midrule
\rowcolor{gray!30}
\multicolumn{5}{l}{\textbf{GPT-OSS-120B admission and placement}} \\
GPT-OSS-120B MoE; 360GB CPU budget; hybrid 2\% 8-bit Adam; batch 32 requested &
Infeasible &
Peak $\le 70$GB; CPU total 360GB &
116.54B trainable parameters; CPU base 1321.38GiB; shortfall
961.38GiB &
With normal Adam-style CPU states, host memory is the first blocker. Reducing
batch size cannot fix optimizer-state memory. \\
\midrule
GPT-OSS-120B MoE; 360GB CPU budget; full 8-bit Adam states; batch 32 requested &
Infeasible &
Peak $\le 70$GB; CPU total 360GB &
CPU base 683.41GiB; shortfall 323.41GiB; no schedule admitted &
Full 8-bit optimizer state helps substantially, but 360GB host memory is still
not enough for this 120B configuration. \\
\midrule
GPT-OSS-120B MoE; 2TiB CPU, 4GB runtime reserve, hybrid 2\% 8-bit Adam; batch 32 &
Optimal solve &
Peak $\le 70$GB; CPU total 2048GB &
Static GPU 41.98GiB; activation budget 28.02GiB; CPU base 1321.38GiB; 37
checkpoints: all GPU activation; objective 0.00037; activation exposed 0ms;
mandatory exposed 31259.54ms &
Once CPU memory is relaxed, activation placement is easy; the bottleneck moves
to unavoidable parameter/gradient streaming. \\
\midrule
GPT-OSS-120B MoE; 2TiB CPU, 4GB runtime reserve, full 8-bit Adam states; batch 32 &
Optimal solve &
Peak $\le 70$GB; CPU total 2048GB &
Static GPU 41.98GiB; activation budget 28.02GiB; CPU base 683.41GiB; 37
checkpoints: all GPU activation; objective 0.00037; activation exposed 0ms;
mandatory exposed 31259.54ms &
8-bit states mainly reduce CPU memory; they do not remove the large
parameter/gradient traffic that must be overlapped. \\
\bottomrule
\end{tabularx}
\end{table*}

\subsection{Qwen3.6-27B Placement View}

The final 27B schedule can be read as a boundary-level placement map:
\begin{itemize}
    \item GPU checkpoints: 0, 10, 15, 16, 20, 22--24, 28--29, 32, 36,
    39--44, 48, 50--52, 56--60, 62--64.
    \item CPU-offloaded checkpoints: 1--5, 7--9, 11--14, 26--27, 30--31,
    33--35, 38, 49.
    \item NVMe-offloaded checkpoint: 21.
    \item Recomputed boundaries: 6, 17--19, 25, 37, 45--47, 53--55, 61.
\end{itemize}

In the unrestricted MILP, each interior layer boundary has four candidate
actions: retain the activation in HBM, offload it to CPU DRAM, offload it to
NVMe, or omit the checkpoint and recompute it during the backward pass. The
final 27B run uses a restricted instance: fixed reference edges require
\(z_{21}=1\), and the fixed reference GPU set requires \(g_{21}=0\). Thus,
boundary 21 is optimized only between CPU and NVMe placement.
Algorithm~\ref{alg:branch_cut} summarizes the general search procedure.

\begin{table*}[t]
\centering
\small
\setlength{\tabcolsep}{4pt}
\caption{Boundary-21 decision trace for the final 27B schedule.
Byte counts are decimal GB because they come directly from bandwidth windows.}
\label{tab:boundary21_example}
\begin{tabularx}{\textwidth}{@{}p{0.18\textwidth}p{0.26\textwidth}X@{}}
\toprule
\textbf{Stage} & \textbf{Decision} & \textbf{Interpretation} \\
\midrule
\rowcolor{gray!30}
\multicolumn{3}{l}{\textbf{Candidate placements}} \\
General MILP option &
$g_{21}=1$ &
Keep boundary 21 in GPU HBM\@. This option is excluded from the final restricted
solve because the reference GPU-checkpoint set fixes $g_{21}=0$. \\
Final-solve candidate &
$c_{21}=1$ &
Offload boundary 21 to CPU; consumes DRAM and PCIe D2H/H2D windows. \\
Final-solve candidate &
$n_{21}=1$ &
Offload boundary 21 to NVMe; consumes SSD capacity, GPU--NVMe path, and
PCIe/GDS windows. \\
General MILP option &
$z_{21}=0$ &
Omit boundary 21 and recompute across it. This option is excluded from the final
restricted solve because the fixed adjacent reference edges require
$z_{21}=1$. \\
\midrule
\rowcolor{gray!30}
\multicolumn{3}{l}{\textbf{Solver-selected assignments}} \\
Selected placement &
$n_{21}=1$; neighbors 20 and 22 stay on GPU &
The solver selects NVMe for boundary 21 while preserving edges $(20,21)$ and
$(21,22)$. \\
Selected read assignment &
NVMe read: 0.198, 0.198, 0.0758, 0.283GB across windows 21, 22, 23, 63 &
The activation is read back in pieces so the SSD endpoint is used inside
available compute windows. \\
Selected write assignment &
NVMe write: 0.2345, 0.2345, 0.0515, 0.2345GB across windows 21, 22, 23, 41 &
The activation is written out in pieces; the solver reports 0ms exposed NVMe
communication. \\
\bottomrule
\end{tabularx}
\end{table*}

\subsection{MoE Note for GPT-OSS-120B}

GPT-OSS-120B is a mixture-of-experts model: only a subset of experts is active
for each token during the forward pass. This reduces active forward compute
relative to a dense model with the same total parameter count. However, full
fine-tuning still needs storage and optimizer states for all trainable expert
parameters. The current \method scheduler is layer-level: it treats the MoE
block as one layer-level object for placement and streaming. Expert-level
placement, expert-aware optimizer-state tiering, and expert-aware communication
overlap are natural follow-up work.

The main text visualizes a separate Qwen3.6-27B time-limit incumbent with an
8\,GB CPU activation budget in Figure~\ref{fig:qwen_json_schedule_map}; it is
not the final 15\,GB-budget schedule summarized above. Additional solver case
studies appear in Section~\ref{sec:app_solver_case_study}, with a direct
GPT-OSS-120B JSON case in Section~\ref{sec:app_gpt_json_case_study}.

\section{MILP Solution Procedure}
\label{sec:app_milp_solution}

\method instantiates the activation-placement and recomputation planner as a
mixed-integer linear program (MILP). PySCIPOpt is used as the modeling interface
and SCIP is used as the underlying solver. For the linear mixed-integer model in
this work, SCIP's exact search is appropriately described as
\emph{branch-and-cut}: a branch-and-bound tree over integer scheduling
decisions, strengthened by presolve, LP relaxations, cutting planes, primal
heuristics, and domain propagation.

\subsection{Canonical MILP Form}

Let \(\mathcal{B}=\{0,\ldots,L\}\) denote activation boundaries and
\(\mathcal{E}\subseteq\{(s,e):0\le s<e\le L\}\) denote legal recomputation
segments. We collect the decision variables from
Eq.~\eqref{eq:lazy_objective} into
\[
y=(x,z,g,c,n,d,h,q,r,\epsilon),
\]
where \(x_{s,e}\) selects recomputation segment \((s,e)\), \(z_i\) indicates
whether boundary \(i\) is materialized, \(g_i,c_i,n_i\) place boundary \(i\) on
GPU HBM, CPU DRAM, or NVMe, \(d/h\) assign CPU offload traffic to D2H/H2D
windows, \(q/r\) assign NVMe write/read traffic, and \(\epsilon\) records
communication not hidden by compute windows. The resulting optimization problem
has the standard MILP form:
\begin{equation}
\label{eq:canonical_milp}
\begin{aligned}
\min_y \quad & a^\top y \\
\text{s.t.}\quad
& A y \le b,\qquad E y = e, \\
& \ell \le y \le u,\\
& x_{s,e}, z_i, g_i, c_i, n_i\in\{0,1\},\\
& d,h,q,r,\epsilon \ge 0.
\end{aligned}
\end{equation}
Here the equality matrix \(E\) contains the checkpoint-path flow constraints
and the unique-placement constraints. The inequality matrix \(A\) contains
GPU/CPU/NVMe capacity constraints and per-window PCIe/NVMe bandwidth
constraints. The linear objective \(a^\top y\) is the weighted sum of
recomputation cost, activation-induced communication exposure, and small
tie-breaking terms. As in Eq.~\eqref{eq:lazy_objective}, mandatory PCIe exposure
that exists without activation offloading is subtracted as a constant baseline.

\paragraph{Relation to the runtime.}
The MILP is solved before training starts. Its output is a serialized schedule
that the layer-streaming runtime consumes directly. Thus the solver is not on the
critical path of every training step; it is an offline planning phase whose
result fixes checkpoint boundaries, storage tiers, and communication windows.

\subsection{Branch-and-Cut Search}

At each node \(v\) of the search tree, SCIP solves the LP relaxation of
Eq.~\eqref{eq:canonical_milp}, replacing binary restrictions by interval bounds
\(0\le x,z,g,c,n\le 1\) and adding any branching decisions inherited from node
\(v\). For a minimization problem, this relaxation gives a valid lower bound
\(\theta(v)\) on all integer schedules in that subtree. If \(\theta(v)\) is
already worse than the best known feasible integer solution, i.e., the
incumbent upper bound, the whole subtree is pruned.

Cutting planes tighten the LP relaxation. A cut is a valid inequality
\(\alpha^\top y\le \beta\) that is satisfied by every feasible integer schedule
but violated by the current fractional LP solution. Adding such cuts improves
the bound without removing any legal \method schedule. If the tightened LP
solution is still fractional, SCIP branches on one fractional integer variable,
for example a placement variable \(n_{21}\) or an edge variable \(x_{20,21}\),
creating child subproblems with that variable fixed to 0 or 1.

\begin{algorithm}[htbp]
\footnotesize
\caption{SCIP Branch-and-Cut Search for the \method MILP}
\label{alg:branch_cut}
\begin{algorithmic}[1]
\Require MILP \(\mathcal{M}\) from Algorithm~\ref{alg:lazy_milp_build}
\Ensure Best feasible schedule \(y^\star\) and final optimality gap, or
infeasibility status
\State Presolve \(\mathcal{M}\) to tighten bounds, remove redundant variables
and constraints, and detect immediate infeasibility.
\State Initialize incumbent upper bound \(U\leftarrow+\infty\) and global lower
bound \(L\leftarrow-\infty\).
\State Insert the root LP relaxation into the node queue.
\While{the node queue is not empty and the stop criterion is not met}
    \State Select a node \(v\) and solve its LP relaxation.
    \If{the LP is infeasible}
        \State Prune \(v\).
    \ElsIf{the node lower bound \(\theta(v)\ge U\)}
        \State Prune \(v\) by bound dominance.
    \Else
        \State Separate valid cutting planes and reoptimize the tightened LP.
        \If{the LP solution is integer feasible}
            \State Update \(y^\star\) and \(U\) if its objective improves the
            incumbent.
        \Else
            \State Select a fractional binary scheduling variable and branch,
            creating two child nodes with that variable fixed to 0 and 1.
        \EndIf
    \EndIf
    \State Update the global lower bound \(L\) from all open nodes.
\EndWhile
\State \Return incumbent schedule \(y^\star\) and gap
\(|U-L|/\max\{1,|U|\}\).
\end{algorithmic}
\end{algorithm}

\paragraph{Why this matters for \method.}
The branch-and-cut procedure is not a heuristic local search over checkpoint
intervals. It systematically searches over checkpoint topology, activation tier
placement, recomputation decisions, and communication-window assignments under
hard memory and bandwidth constraints. The fixed \megatrain heuristic can be
represented as a feasible point in this space; \method asks SCIP to find a
better feasible point and, when solved to optimality, to certify that no better
point exists within the encoded model and solver tolerances. Individual runs may
fix selected variables to a reference schedule. In the final Qwen3.6-27B run,
the optimality certificate applies only after fixing the edges and
GPU-checkpoint set from the \method{} - Hybrid 8-bit variant.

Tables~\ref{tab:appendix_hardware_config}--\ref{tab:appendix_variant_config}
report the resource settings, training protocol, and configuration differences
used to instantiate the MILP and interpret the 27B experiment.

\section{Experimental Configuration}
\label{sec:app_experimental_configuration}

\begin{table*}[t]
\centering
\small
\setlength{\tabcolsep}{4pt}
\caption{Hardware and scheduler resource parameters for the H800 experiments.}
\label{tab:appendix_hardware_config}
\begin{tabularx}{\textwidth}{@{}p{0.26\textwidth}X@{}}
\toprule
\textbf{Item} & \textbf{Configuration} \\
\midrule
Accelerator & NVIDIA H800, 80GB HBM per single-GPU run. \\
Host & Dual Intel Xeon Gold 6448Y CPUs, 128 CPU threads, 2.0TiB DDR5 memory. \\
Local storage & Two local NVMe-backed data mounts, about 7TB each on the
experiment node. \\
GPU--CPU link & PCIe Gen5 x16 node; the measured activation H2D/D2H rate used by
the \method scheduler is 12GB/s per direction. \\
GPU--NVMe path & GDS/cuFile direct path in the final NVMe-aware run; measured
NVMe read/write endpoint rates are 2.83/3.35GB/s. \\
HBM cap & \method schedules use a configured 70GB peak cap; the observed
NVMe-aware training peak is 68.84GB. \\
\bottomrule
\end{tabularx}
\end{table*}

\begin{table*}[t]
\centering
\small
\setlength{\tabcolsep}{4pt}
\caption{Common Qwen3.6-27B / MetaMathQA training configuration. The evaluation
split is also used for periodic loss monitoring.}
\label{tab:appendix_training_config}
\begin{tabularx}{\textwidth}{@{}p{0.26\textwidth}X@{}}
\toprule
\textbf{Item} & \textbf{Configuration} \\
\midrule
\rowcolor{gray!30}
\multicolumn{2}{l}{\textbf{Model and data}} \\
Model & Qwen3.6-27B; Qwen3.5-style text
architecture~\citep{qwen35blog}: 64 layers, hidden size 5120, 24 attention
heads, 4 KV heads, intermediate size 17408. The scheduler computes 23.68B
trainable parameters from this configuration. \\
Precision and attention & bfloat16 training with PyTorch scaled dot-product
attention (SDPA). \\
Dataset & ModelScope MetaMathQA converted to a 70/30 split: 276,499 training
examples and 118,501 evaluation examples. \\
Prompt format & Qwen3.5 non-thinking template; only response tokens contribute
to the training loss. \\
\rowcolor{gray!30}
\multicolumn{2}{l}{\textbf{Optimization and evaluation}} \\
Batch and length & Per-step batch size 72, sequence length 1024, gradient
accumulation 1. \\
Training length & 3841 optimizer steps, equal to one pass over the
276,499-example training split with batch size 72. \\
Optimizer hyperparameters & Adam-style optimizer with \(\beta_1=0.9\),
\(\beta_2=0.999\), \(\epsilon=10^{-8}\), learning rate \(10^{-5}\), weight
decay 0.01, max grad norm 1.0, seed 42. \\
Evaluation & Initial evaluation on 0.1 of the evaluation split before training
(\(\approx\)11,850 examples), then evaluation on 10,000 sampled examples every
0.1 epoch; evaluation batch size 72, seed 42. Final exact-match evaluation uses
the full evaluation split and the same non-thinking prompt template. \\
Safety checks & Non-finite gradients are skipped; training aborts after three
consecutive non-finite-gradient events. \\
\bottomrule
\end{tabularx}
\end{table*}

\begin{table*}[t]
\centering
\scriptsize
\setlength{\tabcolsep}{3pt}
\caption{Configuration differences among the compared Qwen3.6-27B variants.
All use the same model, data split, batch size, and sequence length; the
\method{} - MILP row is a partial experiment without final accuracy. A minus
sign denotes removal of the named component from the complete system.}
\label{tab:appendix_variant_config}
\begin{tabularx}{\textwidth}{@{}p{0.14\textwidth}p{0.27\textwidth}Xp{0.19\textwidth}@{}}
\toprule
\textbf{Variant} & \textbf{Optimizer} & \textbf{Scheduling policy} &
\textbf{Runtime setting} \\
\midrule
\megatrain baseline & DeepSpeed CPUAdam & \megatrain-compatible fixed checkpoint interval
4; no optimized activation-placement solve. &
Baseline execution setting. \\
\midrule
\method & Hybrid 8-bit operator: DeepSpeed CPUAdam plus 2\% CPU 8-bit AdamW
state slice with block size 4096, together with fast gradient clipping. &
Restricted GPU/CPU/NVMe solve with recomputation edges and GPU checkpoints from
the \method{} - Hybrid 8-bit variant fixed: peak cap 70GB, CPU/NVMe activation
budgets 15/32GB, and final placement counts 30/21/1. &
Final NVMe-aware runtime setting. \\
\midrule
\method{} - MILP & Same complete Hybrid 8-bit operator as the full system. &
MILP scheduling disabled. &
Partial experiment without final accuracy measurement. \\
\midrule
\method{} - Hybrid 8-bit & DeepSpeed CPUAdam; both the CPU 8-bit AdamW state
slice and fast gradient clipping are disabled. &
MILP-selected activation schedule retained with the 70GB peak cap and PCIe
bandwidth constraints. &
Optimizer-side ablation setting. \\
\bottomrule
\end{tabularx}
\end{table*}

\section{Experimental Server Environment}
\label{sec:app_server_environment}

Table~\ref{tab:appendix_server_hardware} and
Table~\ref{tab:appendix_server_software} report the concrete server environment
used for the H800 experiments. These are reproducibility details recorded on
the experiment server. They are separate from the scheduler resource model in
Appendix~\ref{sec:app_experimental_configuration}: that appendix reports the
abstract capacities and bandwidths consumed by the MILP, while this section
reports the actual operating system, devices, drivers, and Python package
versions of the experiment server.

\begin{table*}[t]
\centering
\small
\setlength{\tabcolsep}{4pt}
\caption{Experimental server hardware and system configuration.}
\label{tab:appendix_server_hardware}
\begin{tabularx}{\textwidth}{@{}p{0.27\textwidth}X@{}}
\toprule
\textbf{Item} & \textbf{Measured configuration} \\
\midrule
\rowcolor{gray!30}
\multicolumn{2}{l}{\textbf{Host system}} \\
Operating system & Ubuntu 22.04.5 LTS, Linux kernel 5.15.0-113-generic,
x86\_64. \\
CPU & 2\(\times\) Intel Xeon Gold 6448Y sockets; 32 cores per socket; 2 threads
per core; 128 logical CPUs. The CPU supports AVX-512, AVX-512 BF16, AMX BF16,
and AMX INT8 instructions. \\
System memory & 2.0TiB host DRAM; swap disabled. \\
\rowcolor{gray!30}
\multicolumn{2}{l}{\textbf{Accelerators and interconnect}} \\
GPU devices & 8\(\times\) NVIDIA H800 GPUs. Each GPU reports 81,559MiB HBM
through \texttt{nvidia-smi}, i.e., the 80GB H800 class used for single-GPU
experiments. \\
GPU driver and CUDA & NVIDIA driver 575.57.08; CUDA 12.9 reported by
\texttt{nvidia-smi}. \\
PCIe capability & Each H800 reports PCIe Gen5 x16 as the maximum link
capability. The \method scheduler uses a conservative 12GB/s per-direction
activation-transfer bandwidth for CPU--GPU traffic. \\
GPU interconnect & The eight H800 GPUs are connected by NVLink. The reported
experiments use one GPU and do not rely on inter-GPU communication. \\
\rowcolor{gray!30}
\multicolumn{2}{l}{\textbf{Local storage}} \\
Local storage & Root filesystem: 893.8GB RAID device. Two data mounts are each
backed by a 7TB Samsung MZQL27T6HBLA NVMe SSD\@. The \method NVMe-aware offload
path uses local NVMe storage. \\
\bottomrule
\end{tabularx}
\end{table*}

\begin{table*}[t]
\centering
\small
\setlength{\tabcolsep}{4pt}
\caption{Software stack used by the \method experiments and runtime.}
\label{tab:appendix_server_software}
\begin{tabularx}{\textwidth}{@{}p{0.30\textwidth}X@{}}
\toprule
\textbf{Component} & \textbf{Version / setting} \\
\midrule
Python & Python 3.12.13. \\
PyTorch stack & PyTorch 2.10.0+\texttt{cu129}; CUDA runtime 12.9; cuDNN 91701;
Triton 3.6.0. \\
Training libraries & DeepSpeed 0.19.2; Transformers 5.13.0; Datasets 4.0.0;
W\&B 0.28.0; NumPy 2.5.1. \\
Optimization solver & PySCIPOpt 6.2.1 wrapping SCIP for the MILP scheduling
solve. \\
Auxiliary CUDA/sequence packages & flash-linear-attention 0.5.1
(\texttt{fla}); causal-conv1d 1.6.2.post1. \\
Compilation tools & GCC 11.4.0; NVIDIA \texttt{nvcc} 12.9.41. \\
\bottomrule
\end{tabularx}
\end{table*}

%% file: secs/11_reproducibility.tex
\section{Reproducibility Notes}
\label{sec:app_reproducibility}

\parahead{Training protocol}
The main 27B experiment fine-tunes Qwen3.6-27B on MetaMathQA with a 70/30
training/evaluation split, sequence length 1024, batch size 72, and one epoch.
At this batch size, the training split contains 3841 optimizer steps. The
protocol runs an initial evaluation and then evaluates every 0.1 epoch on
10{,}000 sampled evaluation examples; the reported final accuracy uses the full
evaluation split for the final \method configuration.

\parahead{Hardware checks}
The server has eight NVIDIA H800 80GB GPUs, two Intel Xeon Gold 6448Y sockets,
128 CPU threads, 2.0\,TiB host memory, and two local NVMe-backed data mounts;
each reported run uses one H800. GPU memory and PCIe Gen5 x16 were confirmed
with \texttt{nvidia-smi}; CPU and storage capacities were confirmed with
standard system inspection tools. The NVMe read/write endpoint values used by
the scheduler are 2.83/3.35\,GB/s. Full hardware and software configurations
appear in Tables~\ref{tab:appendix_server_hardware}
and~\ref{tab:appendix_server_software}.

\parahead{Reproducibility package}
The accompanying release contains the scheduler implementation, training
configurations, evaluation scripts, and solved schedules used by the reported
setup. It is available at uploaded source code.

%% file: secs/12_impact.tex
\section{Broader Impact and Compute Resources}
\label{sec:app_broader_impact}

\parahead{Broader Impact}
The intended benefit of \method is to make large-model fine-tuning more
accessible to groups with limited accelerator memory. Lowering the hardware
barrier can help academic labs and smaller organizations run controlled
experiments without relying exclusively on large shared clusters. The same
capability can also reduce the friction of training models for harmful uses, so
deployment should follow the same data governance, safety evaluation, and access
control practices expected for the underlying model and dataset.

\parahead{Compute Resources}
The reported experiments use single H800 80GB and RTX 3090 24GB GPUs. The
primary 27B experiment uses one H800 with CPU DRAM and local NVMe storage. The
paper reports per-run throughput and memory use rather than total project
compute. Reproducing the complete 27B run requires one epoch over the 70\%
MetaMathQA training split at batch size 72 and sequence length 1024,
corresponding to 3841 optimizer steps in the reported protocol.

%% file: secs/13_case_study.tex
\section{Visual Solver Case Studies}
\label{sec:app_solver_case_study}

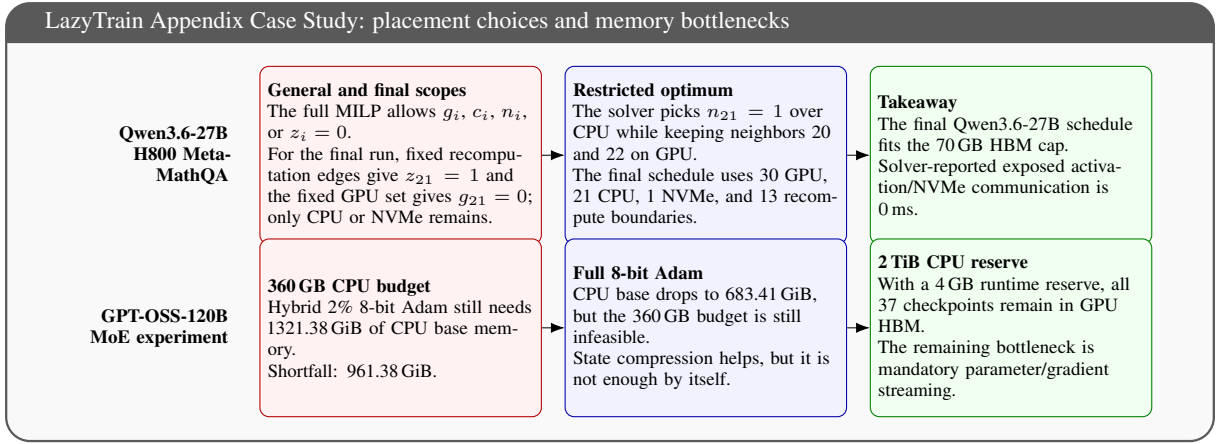
\begin{figure*}[t]
\centering
\small
\begin{tcolorbox}[colback=gray!5!white, colframe=black!65,
title=LazyTrain Appendix Case Study: placement choices and memory bottlenecks,
boxrule=0.3mm, width=\textwidth, arc=3mm, auto outer arc=true]
\centering
\begin{tikzpicture}[
  node distance=0.34cm and 0.30cm,
  bench/.style={font=\bfseries\scriptsize, align=right, text width=0.135\textwidth},
  box/.style={draw, rounded corners=1.2mm, align=left, text width=0.235\textwidth,
    inner sep=3pt, font=\scriptsize, minimum height=2.35cm},
  redbox/.style={box, fill=red!5, draw=red!70!black},
  bluebox/.style={box, fill=blue!5, draw=blue!65!black},
  greenbox/.style={box, fill=green!6, draw=green!50!black},
  arrow/.style={-{Latex[length=1.8mm]}, line width=0.35pt}
]
\node[bench] (qwenlabel) {Qwen3.6-27B\\H800 MetaMathQA};
\node[redbox, right=of qwenlabel] (qwenred) {\textbf{General and final scopes}\\
The full MILP allows $g_i$, $c_i$, $n_i$, or $z_i=0$.\\
For the final run, fixed recomputation edges give $z_{21}=1$ and the fixed GPU set
gives $g_{21}=0$; only CPU or NVMe remains.};
\node[bluebox, right=of qwenred] (qwenblue) {\textbf{Restricted optimum}\\
The solver picks $n_{21}=1$ over CPU while keeping neighbors 20 and 22 on GPU.\\
The final schedule uses 30 GPU, 21 CPU, 1 NVMe, and 13 recompute boundaries.};
\node[greenbox, right=of qwenblue] (qwengreen) {\textbf{Takeaway}\\
The final Qwen3.6-27B schedule fits the 70\,GB HBM cap.\\
Solver-reported exposed activation/NVMe communication is $0$\,ms.};
\draw[arrow] (qwenred) -- (qwenblue);
\draw[arrow] (qwenblue) -- (qwengreen);

\node[bench, below=1.42cm of qwenlabel] (gptlabel) {GPT-OSS-120B\\MoE experiment};
\node[redbox, right=of gptlabel] (gptred) {\textbf{360\,GB CPU budget}\\
Hybrid 2\% 8-bit Adam still needs 1321.38\,GiB of CPU base memory.\\
Shortfall: 961.38\,GiB.};
\node[bluebox, right=of gptred] (gptblue) {\textbf{Full 8-bit Adam}\\
CPU base drops to 683.41\,GiB, but the 360\,GB budget is still infeasible.\\
State compression helps, but it is not enough by itself.};
\node[greenbox, right=of gptblue] (gptgreen) {\textbf{2\,TiB CPU reserve}\\
With a 4\,GB runtime reserve, all 37 checkpoints remain in GPU HBM.\\
The remaining bottleneck is mandatory parameter/gradient streaming.};
\draw[arrow] (gptred) -- (gptblue);
\draw[arrow] (gptblue) -- (gptgreen);
\end{tikzpicture}
\end{tcolorbox}
\caption{Appendix solver case study. Red boxes mark the bottleneck state, blue
boxes mark the solver-selected or strengthened configuration, and green boxes
mark the resulting takeaway. The top row distinguishes the general action space
from the restricted Qwen3.6-27B boundary-21 solve; the bottom row follows the
GPT-OSS-120B host-memory sweep.}
\label{fig:lazytrain_solver_case_study}
\end{figure*}

Table~\ref{tab:appendix_solver_snapshots} and
Table~\ref{tab:boundary21_example} report the detailed solver values.
Figure~\ref{fig:lazytrain_solver_case_study} summarizes the same evidence as a
visual progression: red marks the constrained state, blue the selected or
strengthened configuration, and green the resulting system implication.

The top row makes the boundary-level solver output easier to read: the general
model admits GPU, CPU, NVMe, and recomputation choices, whereas the final run
fixes the recomputation path and GPU-checkpoint set from the
\method{} - Hybrid 8-bit variant before selecting NVMe for boundary 21. In this
restricted experiment, the solver reports all
activation movement hidden inside compute windows. A separate JSON-derived
Qwen3.6-27B time-limit incumbent with a tighter 8\,GB CPU activation budget
appears in Figure~\ref{fig:qwen_json_schedule_map} in the main text. The bottom
row shows that host memory is a distinct constraint from activation placement:
the 1321.38\,GiB hybrid-state and 683.41\,GiB full-8-bit-state requirements both
exceed the 360\,GB budget. A 2\,TiB host budget makes the placement feasible but
leaves mandatory parameter/gradient streaming as the remaining overlap problem.

\subsection{GPT-OSS-120B Solver-Output Case}
\label{sec:app_gpt_json_case_study}

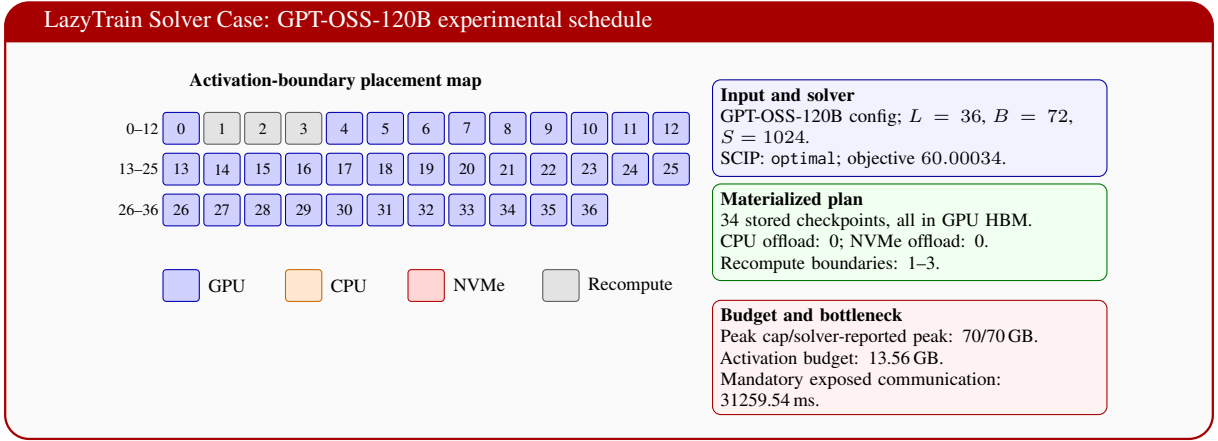
\begin{figure*}[t]
\centering
\small
\begin{tcolorbox}[colback=gray!4!white, colframe=red!65!black,
title=LazyTrain Solver Case: GPT-OSS-120B experimental schedule,
boxrule=0.3mm, width=\textwidth, arc=3mm, auto outer arc=true]
\centering
\begin{tikzpicture}[
  x=0.54cm,
  y=0.54cm,
  tile/.style={draw, rounded corners=0.45mm, minimum width=0.48cm,
    minimum height=0.42cm, inner sep=0pt, font=\tiny},
  note/.style={draw, rounded corners=1.1mm, align=left, inner sep=3pt,
    font=\scriptsize, text width=0.335\textwidth}
]
\tikzset{
  G/.style={fill=blue!18, draw=blue!70!black},
  C/.style={fill=orange!18, draw=orange!80!black},
  N/.style={fill=red!16, draw=red!70!black},
  R/.style={fill=gray!22, draw=gray!70!black}
}
\node[anchor=west, font=\bfseries\scriptsize] at (0,1.15)
  {Activation-boundary placement map};
\node[note, fill=blue!5, draw=blue!60!black, anchor=north west] at (13.0,1.20)
  {\textbf{Input and solver}\\
  GPT-OSS-120B config; \(L=36\), \(B=72\), \(S=1024\).\\
  SCIP: \texttt{optimal}; objective \(60.00034\).};
\node[note, fill=green!6, draw=green!45!black, anchor=north west] at (13.0,-1.35)
  {\textbf{Materialized plan}\\
  34 stored checkpoints, all in GPU HBM.\\
  CPU offload: 0; NVMe offload: 0.\\
  Recompute boundaries: 1--3.};
\node[note, fill=red!5, draw=red!65!black, anchor=north west] at (13.0,-4.20)
  {\textbf{Budget and bottleneck}\\
  Peak cap/solver-reported peak: 70/70\,GB.\\
  Activation budget: 13.56\,GB.\\
  Mandatory exposed communication: 31259.54\,ms.};
\foreach \row/\range in {0/{0--12},1/{13--25},2/{26--36}} {
  \node[anchor=east, font=\tiny] at (-0.35,-\row) {\range};
}
\foreach \i/\s/\row/\col in {
0/G/0/0,1/R/0/1,2/R/0/2,3/R/0/3,4/G/0/4,5/G/0/5,6/G/0/6,
7/G/0/7,8/G/0/8,9/G/0/9,10/G/0/10,11/G/0/11,12/G/0/12,
13/G/1/0,14/G/1/1,15/G/1/2,16/G/1/3,17/G/1/4,18/G/1/5,
19/G/1/6,20/G/1/7,21/G/1/8,22/G/1/9,23/G/1/10,24/G/1/11,25/G/1/12,
26/G/2/0,27/G/2/1,28/G/2/2,29/G/2/3,30/G/2/4,31/G/2/5,
32/G/2/6,33/G/2/7,34/G/2/8,35/G/2/9,36/G/2/10} {
  \node[tile,\s] at (\col,-\row) {\i};
}
\node[tile,G,label=right:{\scriptsize GPU}] at (0,-3.85) {};
\node[tile,C,label=right:{\scriptsize CPU}] at (3.0,-3.85) {};
\node[tile,N,label=right:{\scriptsize NVMe}] at (6.0,-3.85) {};
\node[tile,R,label=right:{\scriptsize Recompute}] at (9.3,-3.85) {};
\end{tikzpicture}
\end{tcolorbox}
\caption{Boundary-wise JSON schedule map for the GPT-OSS-120B solver experiment
at batch size 72. Each tile is one activation boundary from 0 to 36. The solver
keeps every materialized activation checkpoint in GPU HBM and selects
recomputation for boundaries 1--3; the JSON reports 0\,ms exposed activation and
NVMe communication, while mandatory parameter/gradient traffic remains exposed.}
\label{fig:gpt_json_schedule_map}
\end{figure*}

Figure~\ref{fig:gpt_json_schedule_map} renders a second solver-output JSON
for the GPT-OSS-120B Hugging Face configuration at batch size 72. This is a
separate solver experiment from the batch-32 GPT-OSS results in
Table~\ref{tab:appendix_solver_snapshots}: the JSON reports an optimal SCIP
solve, materializes 34 activation boundaries, and recomputes the first three
interior boundaries.

This GPT-OSS-120B case shows a different solver regime from the Qwen3.6-27B
case. Under the 2\,TiB host-memory experimental setting, the activation-placement
decision is simple: no CPU or NVMe activation offload is selected, and the
solver uses only a short recomputation prefix. The objective value primarily
reflects that recomputation. Mandatory parameter/gradient exposure is reported
separately as a diagnostic baseline and is not charged to the placement
objective.